\documentclass[10pt,twocolumn,letterpaper]{article}

\usepackage{cvpr}              

\usepackage{multirow}
\usepackage{graphicx}
\usepackage{amsmath}
\usepackage{amssymb}
\usepackage{booktabs}
\usepackage[accsupp]{axessibility}  
\usepackage{url}
\usepackage{lipsum}

\usepackage[pagebackref,breaklinks,colorlinks]{hyperref}

\usepackage[capitalize]{cleveref}
\crefname{section}{Sec.}{Secs.}
\Crefname{section}{Section}{Sections}
\Crefname{table}{Table}{Tables}
\crefname{table}{Tab.}{Tabs.}

\def\cvprPaperID{5042} 
\def\confName{CVPR}
\def\confYear{2022}

\begin{document}

\title{SVIP: Sequence VerIfication for Procedures in Videos}

\author{Yicheng Qian$^1$, Weixin Luo$^2$, Dongze Lian$^{1,5}$, Xu Tang$^3$, Peilin Zhao$^4$, Shenghua Gao$^{1,6,7\dag}$\\
\\
{}$^{1}$ShanghaiTech University, {}$^{2}$Meituan, {}$^{3}$Xiaohongshu Inc., {}$^{4}$Tencent AI Lab\\
{}$^{5}$National University of Singapore, {}$^{6}$Engineering Research Center of Intelligent Vision and Imaging\\
{}$^{7}$Shanghai Engineering Research Center of Energy Efficient and Custom AI IC\\
{\tt\small \{qianych, luowx, liandz, gaoshh\}@shanghaitech.edu.cn, }\\ {\tt\small tangshen@xiaohongshu.com,  masonzhao@tencent.com}
}
\maketitle


\newcommand \footnoteONLYtext[1]
{
	\let \mybackup \thefootnote
	\let \thefootnote \relax
	\footnotetext{#1}
	\let \thefootnote \mybackup
	\let \mybackup \imareallyundefinedcommand
}
\footnoteONLYtext{\dag ~Corresponding Author}

\begin{abstract}
   In this paper, we propose a novel sequence verification task that aims to distinguish positive video pairs performing the same action sequence from negative ones with step-level transformations but still conducting the same task. Such a challenging task resides in an open-set setting without prior action detection or segmentation that requires event-level or even frame-level annotations. To that end, we carefully reorganize two publicly available action-related datasets with step-procedure-task structure. To fully investigate the effectiveness of any method, we collect a scripted video dataset enumerating all kinds of step-level transformations in chemical experiments. Besides, a novel evaluation metric 
  Weighted Distance Ratio is introduced to ensure equivalence for different step-level transformations during evaluation. In the end, a simple but effective baseline based on the transformer encoder with a novel sequence alignment loss is introduced to better characterize long-term dependency between steps, which outperforms other action recognition methods. Codes and data will be released\footnote{\url{https://github.com/svip-lab/SVIP-Sequence-VerIfication-for-Procedures-in-Videos}}. 
\end{abstract}

%
%
%
%



\begin{figure}[t]
  \centering
  \includegraphics[width=0.95\columnwidth]{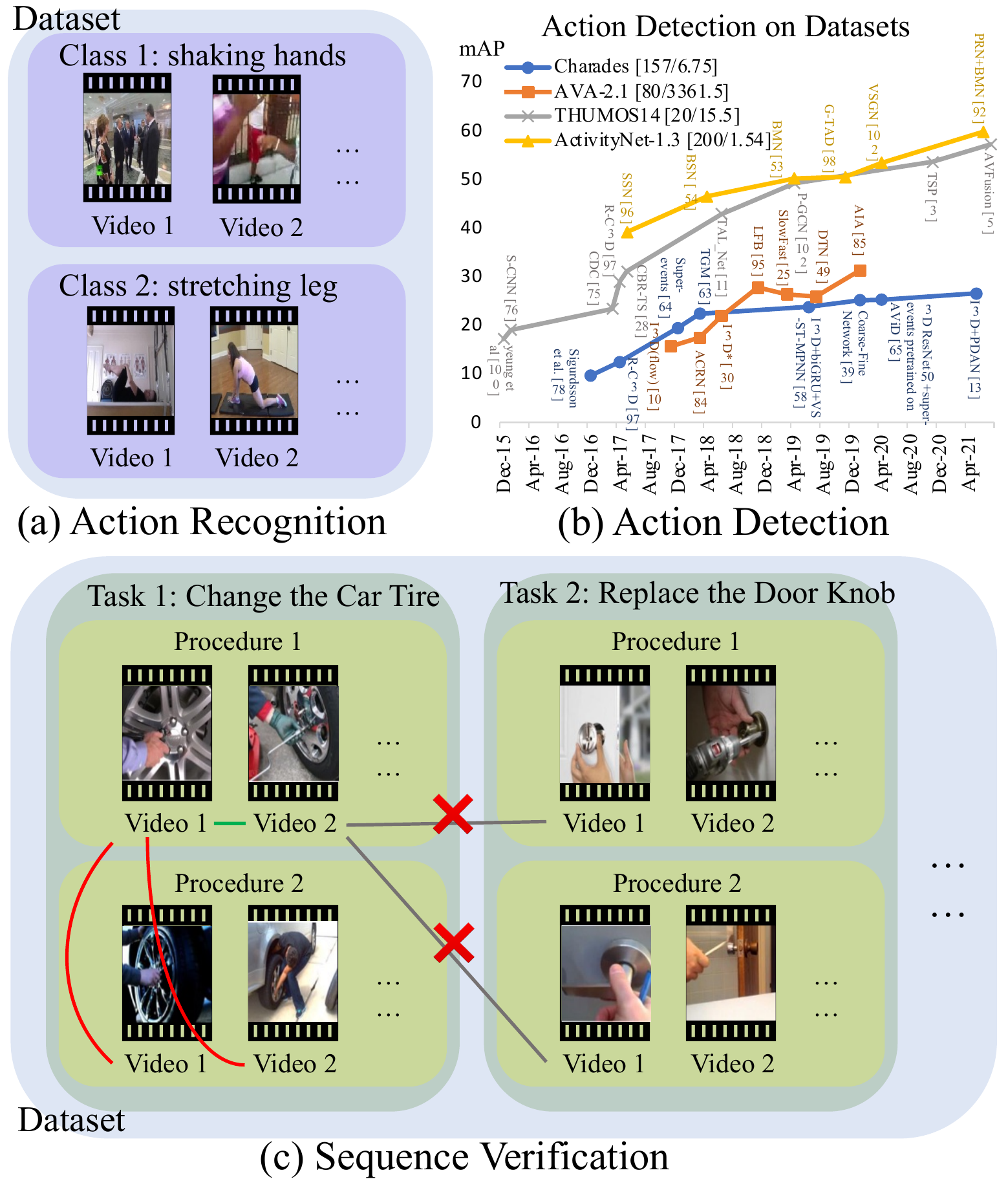}
  \caption{Comparison between traditional action tasks and sequence verification. \textbf{(a)} Action recognition datasets generally consist of various action categories containing videos;  \textbf{(b)} Results of action detection methods on different datasets in recent years. $[a/b]$  means the corresponding dataset contains $a$ action classes and $b$ action instances per video; \textbf{(c)} Sequence verification dataset aims to verify the procedures in the same task. Cross-task verification is dropped due to its simplicity.}
  \label{fig1} 
\end{figure}

\section{Introduction}



In recent years, short-form videos filming people's daily life widely disseminated on social media, which leads to the spurt of activity videos and greatly facilitated the research on video understanding \cite{ryoo2019assemblenet, beery2020context, liu2021video, gu2018ava, tang2019coin, damen2020epic} as well. One can see from these videos that most daily activities are accomplished by serial steps instead of a single step. Such sequential steps form a procedure of which key-steps obey intrinsic consistency, while different participants may accomplish the same activity by different procedures with step-level divergence, as shown in Figure \ref{fig1} (c). In this paper, we advocate a novel action task \textbf{sequence verification} which intends to verify whether the procedures in two videos are step-level consistent, which can be applied to multiple potential tasks such as instructional training and performance scoring. 
To better demonstrate this task, we define related terms specifically. \textit{Step:} a human-action or human-object-interaction atomic unit that is always labeled by a verb, a noun, and even prepositions, \textit{e.g.}, ’remove the old wrapper’; \textit{procedure:} a sequence of steps performed in the chronological order to accomplish a certain goal, \textit{e.g.}, ’remove the old wrapper - wrap with the new wrapper’; \textit{task:} an activity that needs to be accomplished within a defined period of time or by a deadline, \textit{e.g.}, ’Rewrap battery’ and ’Change the car tire’ in COIN. We note that a task can be accomplished by various procedures; \textit{video:} each video performs one procedure of a certain task; \textit{P/N pairs:} two videos performing an identical procedure form a positive pair, while those performing different procedures from the same task form a negative pair.

\emph{Why do we need sequence verification?} Traditional action tasks such as action recognition~\cite{wang2016temporal, lin2019tsm, zhou2018temporal}, action localization~\cite{li2020deep, shou2016temporal, chao2018rethinking} and action segmentation~\cite{lea2017temporal, fayyaz2020sct, chinayi_ASformer} have achieved significant progress due to the development of CNN as well as recently prevalent visual transformer \cite{dosovitskiy2020image}. However, most of these tasks follow a close-set setting with a limitation of predefined categories, illustrated in Figure \ref{fig1} (a). Besides, accurate annotations of steps in numerous videos are extremely time-consuming and labor-intensive, followed by boundary ambiguities that have been studied in recent work~\cite{tang2019coin, zhukov2019cross, shen2021learning, kukleva2019unsupervised} though. However, our proposed sequence verification task circumvents both of these problems by verifying any video pair according to their distance in embedding space. In this way, the sequence verification task neither requires the predefined labels nor consumes the intensive step annotations, which can easily handle the open-set setting.


As shown in Figure \ref{fig1} (c), our proposed sequence verification aims to verify those procedures with semantic-similar steps rather than being associated with totally irrelevant tasks, which enables it to concentrate more on action step association rather than background distinction.  Thus, an appropriate dataset is crucial to perform this task well. However, existing trimmed video datasets such as UCF101 \cite{soomro2012ucf101}, Kinetics \cite{carreira2017quo}, and Moments in Time \cite{monfort2019moments} etc. are leveraged to carry out single-label action recognition. On the other hand, untrimmed video datasets like EPIC-KITCHENS \cite{damen2018scaling}, Breakfast \cite{kuehne12}, Hollywood Extended \cite{bojanowski2014weakly}, ActivityNet \cite{caba2015activitynet} provide videos composed by multiple sub-actions and the corresponding step annotations, but they do not collect videos that especially performs similar or identical procedures. Thus, they cannot be used directly for sequence verification. To this end, we rearrange some datasets such as COIN \cite{tang2019coin} and Diving48 \cite{li2018resound} where each task contains multiple videos recording different procedures, and each video has step-level annotations. Generally, videos with the same procedure are assigned to an individual category for training. Positive pairs and negative ones for testing are collected within the same procedure and cross different procedures in the same task, respectively. It should be noticed that these unscripted videos in the same procedure could be with a large appearance variance due to background divergence and personal preference, which makes sequence verification more challenging. Apart from that, we introduce a scripted filming dataset performing chemical procedures, where it includes all kinds of step-level transformations such as deletions, additions, and order exchanges. Thus, the effectiveness of any algorithm can be well justified by this newly proposed dataset. Additionally, since more step transformations may lead to larger feature distances which is unfair compared to less ones, we introduce a new evaluation metric Weighted Distance Ratio to make sure that every negative pair will be counted equally regardless of its step-level difference during evaluation.

As an unprecedented task, sequence verification may be solved by off-the-shelf action detectors~\cite{sigurdsson2017asynchronous, xu2017r, piergiovanni2018learning, piergiovanni2019temporal, mavroudi2020representation, kahatapitiya2021coarse, piergiovanni2020avid, dai2021pdan, carreira2017quo, sun2018actor, girdhar2018better, wu2019long, feichtenhofer2019slowfast, li2019deformable, tang2020asynchronous}. However, their performance on Charades~\cite{sigurdsson2016hollywood} or AVA~\cite{gu2018ava}, shown in Figure \ref{fig1} (b), is not satisfactory to conduct step-level detection before verification. Although \cite{yeung2016end, shou2016temporal, shou2017cdc, xu2017r, gao2017cascaded, chao2018rethinking, zeng2019graph, alwassel2021tsp, bagchi2021hear, xiong2017pursuit, lin2018bsn, lin2019bmn, xu2020g, zhao2021video, wang2021proposal} perform well on ActivityNet \cite{caba2015activitynet} or THUMOS14 \cite{idrees2017thumos}, it lacks persuasion since the two datasets either contain a few action classes or action instances per video. Thus, we introduce a simple but effective baseline CosAlignment Transformer (abbreviated as CAT), which leverages 2D convolution to extract discriminative features from sampled frames and utilizes a transformer encoder to model inter-step temporal correlation in a video clip. Whereas representing the whole video with multiple steps as a single feature vector may lose information corresponding to the order of steps in a procedure. Thus, we introduce a sequence alignment loss that aligns each step in a positive video pair via the cosine similarities between two videos. The results show that our proposed method significantly outperforms other action recognition methods in the sequence verification task.  

We summarize our contributions as follows:

    i) \textbf{Problem setting:} We propose a new task, sequence verification. To our knowledge, this is the first task focusing on procedure-level verification between videos.  

    ii) \textbf{Benchmark:} We rearrange two unscripted video datasets with significant diversity and propose a new scripted dataset with multiple step-level transformations to support this task. Moreover, a new evaluation metric is introduced especially for this novel task.

    iii) \textbf{Technical contributions:} We propose a simple but effective baseline that contains a transformer encoder to explicitly model the correlations between steps. Besides, a sequence alignment loss is introduced to improve the sensitivity to step disorder and absence. This novel baseline significantly outperforms other action recognition methods.

\begin{table*}[t]
\centering
\begin{tabular}{l|c|c|c|l|l|l}
\hline
Dataset & \# Tasks & \# Videos & \# Steps & \# Procedures & \# Split Videos & \# Split Samples \\
\hline
COIN-SV    & 36    & 2,114   & 130   & 37~/~268~/~285    & 1,221~/~~451~~/~442      & 21,741~/~1,000~/~400  \\
\hline
Diving48-SV & 1     & 16,997  & 24   & 20~/~~20~~/~8     & 6,035~/~7,938~/~3,024      & 50,000~/~1,000~/~400  \\
\hline
CSV       & 14     & 1,940   & 106   & 45~/~~25~~/~-     & 901~~~/~1,039~/~-     & 8,531~~~/~1,000~/~-  \\
\hline
\end{tabular}
\caption{The statistical information of three datasets. It is listed with an order of training, testing, and validation.}
\label{table1}
\end{table*}

\section{Related Work}
\noindent\textbf{Action Tasks.}
Traditional action-related tasks such as action recognition, action detection, and action segmentation have been greatly developed due to the advances in CNNs. i) As a means of general video representation,  deep-learning-based \textbf{action recognition} can be generally summarized to stream-based methods \cite{simonyan2014two, donahue2015long, feichtenhofer2016convolutional, wang2016temporal, christoph2016spatiotemporal, tran2015learning,qiu2017learning,diba2017temporal,tran2017convnet,carreira2017quo, lin2019tsm, zhou2018temporal} and skeleton-based \cite{du2015hierarchical,weng2017spatio,yan2018spatial, song2016end} methods. Both kinds of methods aim to
produce a feature representation for each trimmed video, to which a video-level label over predefined action categories is predicted according. ii) To seek the interested sub-actions in untrimmed videos, \textbf{action detection}~\cite{lin2019bmn, lin2018bsn, zhao2017temporal, feichtenhofer2019slowfast, sun2018actor, dai2021pdan, piergiovanni2020avid, kahatapitiya2021coarse, mavroudi2020representation, xu2017r, piergiovanni2019temporal, piergiovanni2018learning} are proposed to detect the start and end of sub-action instances and predict their categories. iii) To conduct dense action predictions in untrimmed videos, \textbf{action segmentation} is designed to label each frame including background in videos. With dense annotations, fully-supervised methods \cite{rohrbach2012database, shi2008discriminative, yeung2016end, shou2016temporal, richard2016temporal, lea2016segmental, farha2019ms, miech2020end} rely on sliding windows, Markov models, or temporal convolutional networks to model the temporal relations. However, dense annotations in videos require expensive human-labors as well as consume much time, though the weakly-supervised methods \cite{bojanowski2014weakly, kuehne2017weakly, huang2016connectionist, ding2018weakly, richard2017weakly} with order labels of actions only have achieved satisfactory performance. Last but not least, it still remains a concern if these action-related tasks are able to generalize well to unknown classes in the wild. Different from them, our task has no restriction during inference, so it can easily tackle an open-set setting.

\noindent\textbf{Video datasets.}
Multiple existing video datasets\cite{kuehne2011hmdb,soomro2012ucf101,karpathy2014large,caba2015activitynet,abu2016youtube,kay2017kinetics,murray2012ava,goyal2017something,damen2018scaling,sigurdsson2016hollywood} have being dominated on video understanding during a long period. To begin with, HMDB51 \cite{kuehne2011hmdb} and UCF101 \cite{soomro2012ucf101} that contains 51 and 101 classes of actions, respectively, are introduced for action recognition. Next, Something-Something \cite{goyal2017something} collects 147 classes of interactions between humans and objects in daily life. In addition, ActivityNet \cite{caba2015activitynet} and Kinetics \cite{kay2017kinetics} collect videos from YouTube and builds large-scale action recognition datasets. Other datasets for instructional video summarization and analysis \cite{tang2019coin,rohrbach2012database,song2015tvsum} that are annotated with texts and temporal boundaries of a series of steps, contributes to the understanding of language and vision. EPIC-KITCHENS dataset \cite{damen2018scaling} collects the human actions such as washing glass or cutting bell pepper in kitchen scenes and targets at the first-person perspective to reflect people's goals and motivations. 

\noindent\textbf{Instructional videos analysis.}
Instructional videos are generally accompanied with explanations such as audio or narrations matching the timestamps of sequential actions, which has attracted the research interest in the video understanding community. For instance, step localization \cite{zhou2018towards, tang2019coin, miech2020end} as well as action segmentation \cite{tang2019coin, fried2020learning, ghoddoosian2021hierarchical, piergiovanni2021unsupervised} in instructional videos have been widely studied in the early stage. With the growing attention paid to this research topic, various kinds of tasks related to instructional videos have been proposed, \textit{e.g.}, video captioning \cite{zhou2018towards, luo2020univl, huang2020multimodal, tang2021decembert} which generates the description of a video based on the actions and events, visual grounding \cite{sigurdsson2020visual, huang2018finding} which locates the target in an image according to the language description, and procedure learning \cite{sener2015unsupervised, alayrac2016unsupervised, zhou2018towards, elhamifar2019unsupervised, fried2020learning, shen2021learning} which extracts key-steps.


\begin{figure}[t]
  \centering
  \includegraphics[width=0.95\columnwidth]{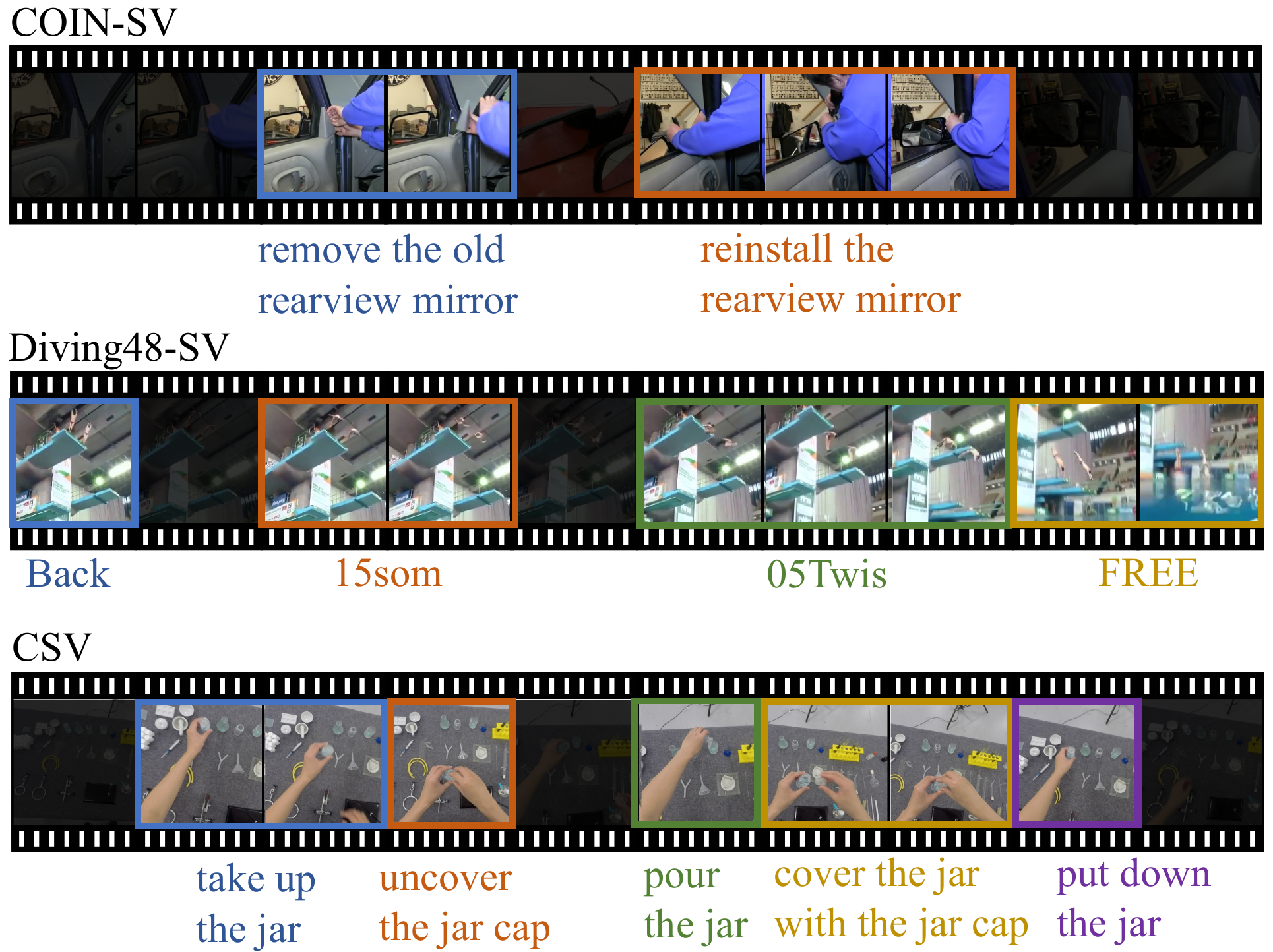}
  \caption{Dataset illustration. COIN-SV contains 36 daily life tasks such as 'Replace rearview mirror glass', 'Make burger', leading to high background diversity. Diving48-SV and CSV consist of videos in the diving competition and chemical experiments scene, respectively. Every video in three datasets is categorized by the sequence of steps it performs, as known as the procedure in the video. Note that only COIN-SV provides temporal annotation for steps, while Diving48-SV and CSV only provide the procedure-level annotation for each video.}
  \label{fig2} 
  
\end{figure}

\section{Data Preparation}\label{sec3}

Due to the intrinsic step-procedure-task structure in the publicly available datasets COIN \cite{tang2019coin}, and Diving48 \cite{li2018resound}, we reorganize these two datasets to support our proposed sequence verification task focusing on verifying various step-level transformations. However, the procedures in the same task may lack enough diversity in these datasets to fully verify the effectiveness of our proposed method for sequence verification. Thus, we collect a novel scripted dataset, Chemical Sequence Verification, enumerating all kinds of procedures in the same task, which will be introduced later. The statistics of these three datasets can be found in Table \ref{table1}. We visualize some samples in Figure \ref{fig2}.

The rest of this section introduces the common structure, specific processing and basic information of these datasets.

\subsection{Common Structure}
As shown in Figure \ref{fig1} (c), each dataset used in this paper contains videos completing various tasks, \textit{e.g.}, the original COIN dataset contains 180 tasks common in daily life. In practice, each individual task can be accomplished by different procedures of which steps as atomic actions still obey certain orders. Meanwhile, the steps of two procedures with different task-orientations will not overlap each other at most times. Thus, we will not introduce sequence verification cross tasks to ensure the challenge.

\subsection{COIN-SV}
COIN \cite{tang2019coin} is a comprehensive instructional video dataset that contains 180 tasks such as 'Replace the door knob', 'Change the car tire', and 'Install a ceiling fan'. This recently proposed dataset is quite challenging due to its background diversity and even significant distinctions between videos of the same procedures, which benefits our proposed sequence verification task. In total, it contains 11827 videos over 4715 procedures, which means COIN is followed by a long-tail distribution where most procedures have one or two videos only. To facilitate the classifier training, we preserve 36 tasks that contains at least one procedure with more than 20 videos and discard the other tasks. Procedures with more than 20 videos are used for training, and the rest are assigned to the validation and testing sets randomly. Since the original split in this dataset is reorganized, we name it COIN-SV.



\subsection{Diving48-SV}
Diving48 \cite{li2018resound} dataset records diving competition videos with 48 kinds of diving procedures standardized by the international federation FINA, which consists of around 18,000 trimmed videos. Each diving procedure is a sub-action sequence of one-step takeoff, two-step movements in flight, and one-step entry. In total, 16997 videos over 48 procedures are publicly available up to now. Obviously, this dataset is less challenging than COIN due to its dual background including a board, a pool, and spectators and less step-level divergence. We assign 20, 8, 20 procedures for the training, validation, and testing sets, respectively. Similar to COIN-SV, we name it Diving48-SV.


\subsection{Chemical Sequence Verification}
Since the videos in COIN-SV and Diving48-SV are gathered from the internet, it is difficult to include all kinds of step-level transformations without predefined scripts, which is crucial for the sequence verification task. To this end, we collect a new dataset named Chemical Sequence Verification (CSV) containing videos with all kinds of step-level transformations such as deletions, additions, and order exchanges. Concretely, volunteers from an egocentric perspective are asked to conduct chemical experiments with predefined scripts. In a word, the CSV dataset includes 14 tasks, and each consists of 5 procedures. We select 45 procedures for training and 25 procedures for testing. CSV has no validation set due to its limited number of procedures/videos. Data gathering process, video annotations, and statistics information are available in the supplementary material.

\begin{figure*}
  \centering
  \includegraphics[width=1.85\columnwidth]{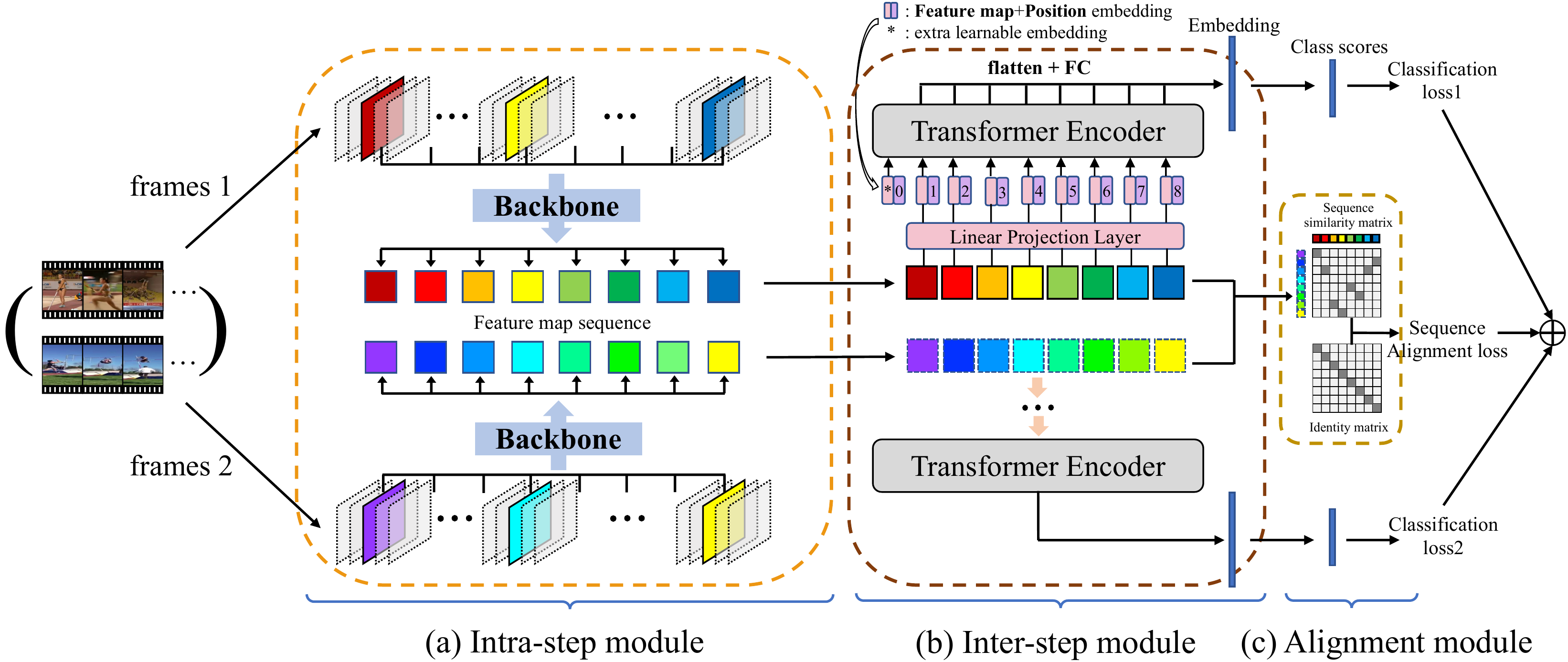}
  \caption{Pipeline overview. \textbf{a)} Applying a 2D backbone on the sampled frames (colored in the input frames) to capture features of individual steps, which we called intra-step features. The output of this module is a sequence of feature maps that are temporally modeled from the sampled frames. \textbf{b)} Applying a transformer encoder to aggregate the sequential feature maps.
  \textbf{c)} This module aims at imposing the two sequences of feature maps to be matched.}
  \label{fig3} 
\end{figure*}

\section{Method}

Classical models for video action recognition \cite{simonyan2014two, donahue2015long, feichtenhofer2016convolutional, wang2016temporal, christoph2016spatiotemporal,tran2015learning,qiu2017learning,diba2017temporal,tran2017convnet,carreira2017quo} aim to predict action categories without paying attention to sub-action orders as many as possible due to simple frame feature aggregation such as pooling. Nevertheless, our task intends to verify two videos with large as well as subtle step-level transformations. For instance, a video performing A, then B, and finally C is treated as a negative sample to another video carrying out A and finally C, while both of them may be successfully predicted by a traditional action classifier. To fit our proposed task, we introduce two remedies over the traditional action classification during training: i) procedures rather than tasks in the training set are regarded as training classes, in order to enable the model to distinguish those procedures even with tiny step-level transformations in the same task; ii) since pooling over frame features may bring order insensitivity to the model, we remain the temporal dimension without any down-sampling operation and it is finally reshaped to the channel dimension, followed by a fully-connected layer with order-sensitivity.


\subsection{Preliminary}\label{sec4.1}
For a certain dataset $D=\{(V_i, S_i)\}_{i=1}^n$, a set of $n$ video clips $V$ are given with corresponding procedure annotations $S$. \textbf{Here we do not use the timestamp annotations of steps since action detectors will not be used in this paper.} We denote the model as $f:\mathbb{R}^{3\times H\times W\times K}\rightarrow\mathbb{R}^C$. $K$ is the number of sampled frames in a video. $H$ and $W$ are frames' height and width, respectively. $C$ is the total number of procedures in the training set. Following the paradigm in face verification \cite{amos2016openface, schroff2015facenet, deng2019retinaface}, we treat sequence verification as a multi-category classification task during training, and videos performing the same procedure are classified into the same category. In the testing phase, we collect the videos from the same procedure to form positive pairs and the videos from different procedures but still in the same task to form negative pairs. Then embedding distance between two videos in a pair indicates the verification score of this pair. The procedure classification loss $L_\text{cls}$ is as follows.

\begin{equation}
L_\text{cls}= \sum_{i=1}^n\delta(f(V_{i}), Y_{i})
\end{equation}
where $\delta$ is the cross-entropy function, $Y_i$ is the $C$-dim one-hot vector whose entry corresponding to $S_i$ is 1.



\subsection{Baseline}
    We utilize a ResNet~\cite{he2016deep} backbone followed by a fully-connected layer to aggregate temporal information and a softmax classification layer as our baseline. Following TSN \cite{wang2016temporal}, we divide each input video into $K$ segments ($K=16$ in our experiments). One frame in each segment is randomly selected to form the input tensor $x\in\mathbb{R}^{3\times H\times W\times K}$, which is fed into the backbone and outputs a tensor with a shape of $D\times K$ where $D$ is feature dimension. Then it is flattened into a vector for order-sensitivity. Finally, a procedure classifier with $C$ categories is appended for training.

\subsection{Transformer Encoder}
    Transformer \cite{vaswani2017attention} has achieved great success in Natural Language Process and it has been applied in multiple computer vision tasks such as image recognition \cite{dosovitskiy2020image, wu2020visual} and object detection \cite{carion2020end,zhu2020deformable}. To better characterize inter-step correlations, we follow \cite{dosovitskiy2020image} and integrate the transformer encoder into the backbone by replacing the global average pooling. As Figure \ref{fig3} describes, we firstly flatten the spatial feature maps and apply a trainable linear projection layer on these flattened vectors, resulting in feature vectors $\mathbf{E}_i\in\mathbb{R}^D~,i=1,2,\cdots, K$. 
    Further, randomly initialized position embedding is added to retain order information. In conclusion, the input $\mathbf{I}$ of a standard transformer encoder is 
    \begin{small}
    \begin{align}\label{encoder input}
    	\textbf{I}=[\mathbf{E}_1; \mathbf{E}_2; \cdots; \mathbf{E}_K]+\mathbf{E}_{pos}, \in\mathbb{R}^{K\times D}.
    \end{align}
    \end{small}
    
    The output $\textbf{O}\in \mathbb{R}^{K\times 1024}$ is flattened and fed into a fully-connected layer for a global representation of the input video. We adopt the sequential features instead of the CLS token since the former explicitly remains order information.



\subsection{Sequence Alignment}
    So far, our proposed method aims to extract a global video representation supervised by the procedure classifier. However, the step order in a procedure is especially important in sequence verification. To make sure two positive procedures are step-level consistent, we propose a Sequence Alignment loss that explicitly imposes feature consistency step-by-step. Specifically, we extract the last spatial feature maps in the backbone and use global average pooling to produce feature vectors for all the frames in a given positive pair $(seq_1,~seq_2)$, where $seq_i$ is the frame sequence sampled from video $i$. Then cosine similarity is calculated for all the frame pairs formed by the the two sequences, resulting in a correlation matrix:

    \begin{equation}
    	\text{corr}_{ij}=\frac{f_{1i}}{\|f_{1i}\|}\cdot\frac{f_{2j}^T}{\|f_{2j}\|},
    \end{equation}
    where $\text{corr}_{ij}$ denotes the similarity value at the $i$-th row and the $j$-th column of the matrix $\text{corr}$, while $f_{1i}$ and $f_{2j}$ represent the $i$-th feature of $seq_1$ and $j$-th feature of $seq_2$, respectively. Next, we perform a softmax function on each row of the similarity matrix to produce $\text{corr}^1$, whose $i$-th row is composed of cosine similarities between the $i$-th feature of $seq_1$ and every feature of $seq_2$. Similarly, we perform a softmax function on each column of the similarity and produce $\text{corr}^2$. We average these two matrices and denote the result as $\text{corr}_{avg}$. The diagonal values of $\text{corr}_{avg}$ are then expected to be close to 1 while other values are expected to be close to 0, since both $\text{corr}^1$ and $\text{corr}^2$ have been normalized by softmax. In other words, we impose two videos in a positive pair to be similar in the feature space frame-by-frame, to some extent step-by-step. Mathematically, our proposed Sequence Alignment loss $L_\text{seq}$ can be defined as:
    \begin{equation}
    \centering
    	L_\text{seq} = \|\mathbf{1}-h(\frac{\text{corr}^1+\text{corr}^2}{2})\|_1,
    \end{equation}
    where $\mathbf{1}$ is a vector whose entries are all one and $h$ is a function to extract the diagonal entries of a matrix.

\subsection{Training Loss}
Now, we train the network by the procedure classification loss and the sequence alignment loss in an end-to-end manner. Thus, the total loss $L$ can be summarized:


\begin{equation}
L = L_\text{cls} + \lambda L_\text{seq}
\end{equation}
Here $\lambda$ is a hyper-parameter and it sets to 1 by default.

\subsection{Testing phase}
During inference, the goal of sequence verification is to distinguish positive pairs from negative pairs. We denote each pair as $P_i = (V_{i_1}, V_{i_2})$. The model takes each video in $P_j$ as input and produces one $d$-dimension visual embedding before the classification layer, which is denoted by $f':\mathbb{R}^{K\times H\times W\times 3}\rightarrow\mathbb{R}^{D'}$. Next, we calculate the normalized Euclidean distance between the two procedures in the embedding space and the verification score $y_i$ is defined:
\begin{align}
& d_i = g(f'(V_{i_1}),f'(V_{i_2}))\label{eq6}\\
& y_i = \left\{
    \begin{aligned}
        & 1, & d_i \leq \tau, \\
        & 0, & otherwise.
    \end{aligned}
    \right.
\end{align}

where $g$ is a function that does $l_2$ normalization over two embeddings firstly and then calculates their Euclidean distance, $ \tau$ is a threshold to decide whether the procedures are consistent. $y=1$ means the procedures in two videos are consistent, otherwise inconsistent.

\section{Experiments}

\subsection{Experimental Details}
\noindent\textbf{Datasets and setup.}
We conduct experiments on COIN-SV, Diving48-SV, and CSV. The specific information of each dataset is available in Section \ref{sec3}. Since this novel task is proposed to solve the open-set setting, there exists no procedure-level overlapping among the training, validation, and testing sets. However, step-level overlapping is unavoidable since different procedures can still contain several common steps.

\noindent\textbf{Implementation Details.}
The ResNet-50 we employ is pre-trained on Kinetics-400 \cite{kay2017kinetics} to avoid over-fitting, while the new layers adopt Kaiming uniform initialization \cite{he2015delving}. The experiments are conducted on 4 NVIDIA TITAN RTX GPUs with batch size $16$, a cosine learning rate scheduler with a base learning rate of $0.0001$, and weight decay $0.01$. Adam \cite{kingma2014adam} is used to optimize the whole network. For efficiency, we resize the raw images to $180 \times 320$. We also leverage horizontal flip, cropping, and color jittering for data augmentation. The feature dimension $D'$ before the classifier layer is set to 128 for all experiments.

\begin{table*}[t]
\centering
\begin{tabular}{lll|ll|ll|l}
\hline
& & & \multicolumn{5}{c}{AUC / WDR} \\
\cline{4-8}
Method & Pretrain & \#Param(M) & \multicolumn{2}{c}{COIN-SV} & \multicolumn{2}{c}{Diving48-SV} & \hspace{0.4cm}CSV \\
\cline{4-8}

& & &\hspace{0.6cm}Val &\hspace{0.6cm}Test &\hspace{0.6cm}Val &\hspace{0.6cm}Test &\hspace{0.6cm}Test \\
\hline
Random &  \multicolumn{1}{c}{-} & \hspace{0.7cm}- & 50.00 / \hspace{0.4cm}- & 50.00 / \hspace{0.4cm}- & 50.00 / \hspace{0.4cm}- & 50.00 / \hspace{0.4cm}-   & 50.00 / \hspace{0.4cm}- \\

TSN \cite{wang2016temporal}   & K-400 & \hspace{0.4cm}22.67 & 53.38 / 0.3651 & 47.01 / 0.3999 & 91.00 / 1.0835 & 81.87 / \textbf{0.6707}   & 59.85 / 0.3447 \\
TRN \cite{zhou2018temporal}   & K-400 & \hspace{0.4cm}23.74 & 54.92 / 0.3665 & \textbf{57.19} / 0.3719 & 90.17 / \textbf{1.1438} &80.69 / 0.5876   & 80.32 / \textbf{0.4677} \\
TSM \cite{lin2019tsm}   & K-400 & \hspace{0.4cm}22.67 & 52.12 / 0.2948 & 51.25 / 0.3872 & 89.41 / 1.0035 & 78.19 / 0.5531   & 62.38 / 0.3308 \\
Swin \cite{liu2021video}  & K-400 & \hspace{0.4cm}26.66 & 47.27 / 0.3895 & 43.70 / 0.3495 & 89.35 / 1.1066 & 73.10 / 0.5316   & 54.06 / 0.3141 \\
\textbf{CAT(ours)} & K-400 & \hspace{0.4cm}72.32 & \textbf{56.81} / \textbf{0.4005} & 51.13 / \textbf{0.4098} & \textbf{91.91} / 1.0642 & \textbf{83.11} / 0.6005 & \textbf{83.02} / 0.4193 \\

\hline
\end{tabular}
\caption{Comparison with action recognition methods on the validation and testing set of COIN-SV, Diving48-SV, and CSV dataset.}
\label{table2}
\end{table*}
 
\noindent\textbf{Baselines.}
Since we are the first to introduce the sequence verification task, there are no existing methods that are specially designed for this task. Considering that we learn video representation during training, which is similar to the action recognition task, we compare our proposed method with some advanced action recognition baselines: \textit{Random}, \textit{ TSN \cite{wang2016temporal}}, \textit{ TRN \cite{zhou2018temporal}}, \textit{ TSM \cite{lin2019tsm}}, and \textit{ Video Swin \cite{liu2021video}}.


\noindent\textbf{Evaluation Metrics.} (1) \textbf{AUC.} We adopt the Area Under ROC Curve (abbreviated as AUC) as one of the measurements, which is commonly used to evaluate the performance of face verification. Higher AUC denotes better performance. (2) \textbf{WDR.} It is short for Weighted Distance Ratio. To begin with, we calculate the mean embedding distance per unit Levenshtein distance for negative pairs in order to guarantee the equivalence of each pair during evaluation because larger step-level transformations always lead to larger embedding distance, discussed in Section \ref{sec5.5}. The mean embedding distance over positive pairs is then computed. In the end, we use the ratio between negative distance and positive distance, namely the Weighted Distance Ratio, as an indicator of performance for all methods. Obviously, its higher value means better performance the methods arrive at. Mathematically, we define WDR as:
\begin{equation}
    \text{WDR} = \frac{\sum_{i=1}^{N}wd_i/N}{\sum_{j=1}^{P}d_j/P},
\end{equation}
where $P$ and $N$ are number of positives and negatives, repectively. $d_i$ and $d_j$ can be easily calculated by Equation \ref{eq6}. $wd_i$ is defined as: 
\begin{equation}
    wd_i = \frac{d_i}{ed_i}
\end{equation}
where $ed_i$ represents the text Levenshtein distance of pair $i$. Levenshtein distance, defined as \textit{the minimum number of operations required to transform one string into the other} can be used as a measurement of how different in terms of steps two procedures are. More explanations and evaluations can be found in Section \ref{sec5.5}.


\subsection{Comparison of Different Methods}
The quantitative results of all the methods on the three datasets are shown in Table \ref{table2}. We can find that our proposed CAT exceeds all other baselines in most cases, evaluated on the AUC metric. It is worth noticing that CAT does not achieve the best WDR in all datasets since the best model is selected by the highest AUC in the validation set. Apart from that, AUC on COIN-SV is extremely inferior compared to the other two datasets, which indicates its significant challenge due to its complex background and procedure diversity. Surprisingly, Video Swin Transformer is inferior to other baselines. We conjecture that it suffers from data insufficiency.



\subsection{Ablation Study}

\begin{table}[t]
\centering
\begin{tabular}{l|l|l|ll}
\hline
Dataset                                      & +TE & +SA & AUC (\%) & WDR \\
\hline
\multirow{3}{*}{COIN-SV}    &      &     & 52.31 & 0.3677 \\
                                                      & \checkmark &     & 55.46 & 0.3839 \\
                          & \checkmark & \checkmark & \textbf{56.81} & \textbf{0.4005} \\
                          \cline{1-1} \cline{2-5}
\multirow{3}{*}{Diving48-SV} &      &     & 90.51    & 1.0093    \\
                          & \checkmark &     & 90.91    & 1.0308    \\
                          & \checkmark & \checkmark & \textbf{91.91} & \textbf{1.0642} \\
                          \cline{1-1} \cline{2-5}
\multirow{3}{*}{CSV}       &      &     & 81.97 & \textbf{0.4403} \\
                          & \checkmark &     & 82.07 & 0.4193 \\
                          & \checkmark & \checkmark & \textbf{83.02} & 0.4193 \\
                          \hline
\end{tabular}
\caption{Comparison between different model structures of our method on the testing set of the CSV dataset.}
\label{table3}
\end{table}

\begin{figure}[t]
  \centering
  \includegraphics[width=0.95\columnwidth]{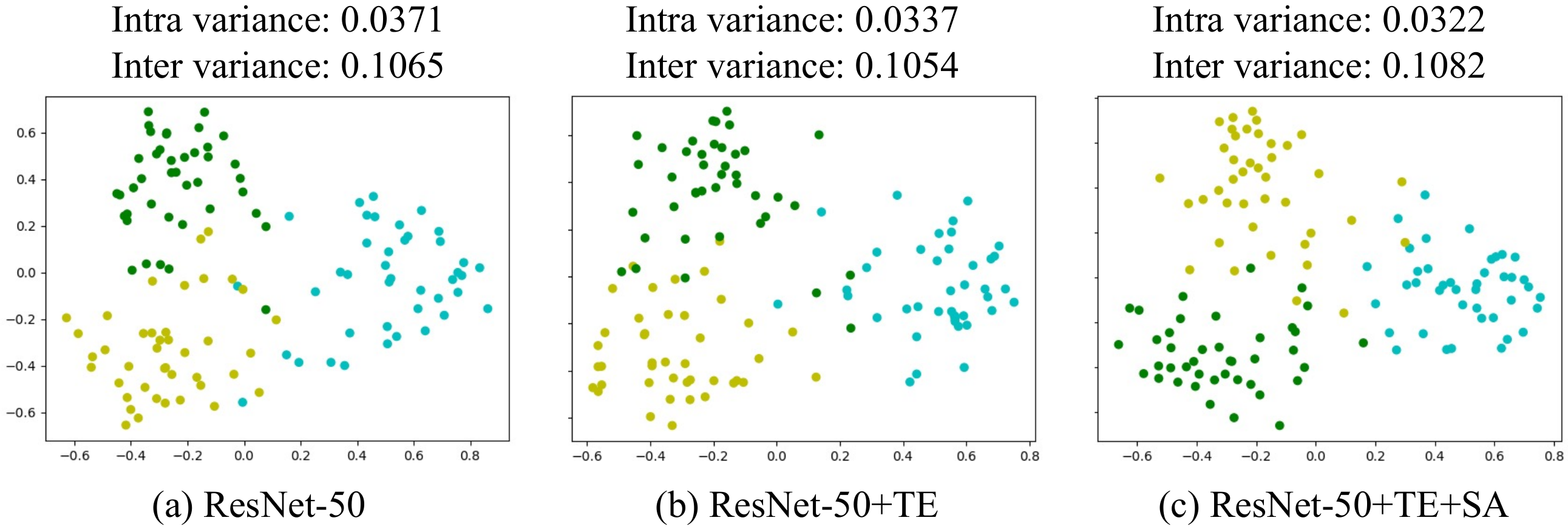}
  \caption{Visualizations of embeddings predicted by different model structures. Three kinds of colored points represent different procedures from the same task. The values above each sub-figure are the averaged intra-procedure variance and the inter-procedure variance, respectively.}
  \label{fig4} 
\end{figure}

In this section, we investigate the effectiveness of the transformer encoder (TE) and sequence alignment (SA) module. The experiments are conducted on the testing set of the CSV dataset if not specially stated. Specifically, we gradually add TE and SA module to the ResNet-50. The results in Table \ref{table3} show that both of the module improve the AUC performance on three datasets. As for WDR, CAT achieves the best performance on COIN-SV and Diving48-SV but it is inferior to the vanilla on CSV.

We also visualize the 128-d embedding vectors extracted by different models via PCA in Figure \ref{fig4}. Concretely, we select the first three procedures in the first task of the CSV dataset. Since they only differ in the order of steps but hold the same step set, we can evaluate the effectiveness of the sequence alignment module for handling the order consistency. As shown, the embeddings extracted by the entire CAT model in sub-figure (c) have the largest inter-procedure variance and the smallest averaged intra-procedure variance. 


\subsection{Performance on Different Splits}
As a reminder, the step-level transformation contains deletions, additions, and order exchange of steps. To verify the order-sensitivity of the sequence alignment module, we further re-divide the testing set of CSV into two splits of which one consists of video pairs containing step additions and deletions, and the other consists of pairs containing order exchange of steps, which refers to \textit{alter-number split} and \textit{alter-order split}, respectively.

The results shown in Table \ref{table4} indicate that CAT without SA module achieves inferior performance on both \textit{alter-number split} and \textit{alter-order split} while introducing the SA module brings more performance gain on \textit{alter-order split}, which strongly supports the motivation of our proposed SA module that enables the model to be more order-sensitive.

\begin{table}[t]
\centering
\begin{tabular}{c|c|c}
\hline
Test Split    & CAT w/o SA                 & CAT w/ SA                           \\
\hline
alter-number  & 73.01 & 75.82 (+2.81)  \\
alter-order   & 80.24 & 86.32 (+6.08) \\
\hline
\end{tabular}
\caption{Results on different test splits. (metric: AUC (\%))}
\label{table4}
\end{table}

\subsection{WDR Curve}\label{sec5.5}

\begin{figure}[t]
  \centering
  \includegraphics[width=0.95\columnwidth]{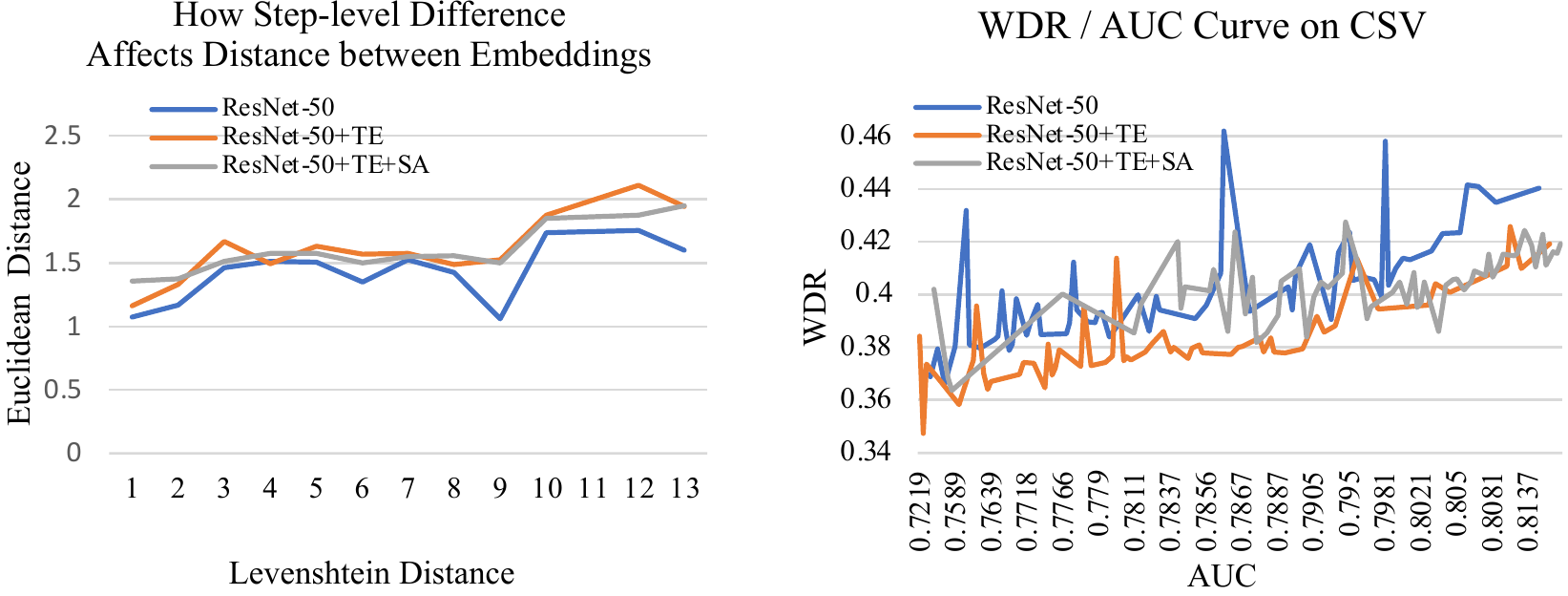}
  \caption{\textbf{Left:} Euclidian Distance between procedures with different degrees of step-level divergence in the embedding space; \textbf{Right:} Curve of WDR versus AUC on the testing split of CSV.}
  \label{fig5} 
\end{figure}

To carefully explore the character of our proposed evaluation metric WDR, we conduct two experiments to study the relationship between embedding distance and Levenshtein distance, and the relationship between WDR and AUC. First of all, the curve in Figure \ref{fig5} \textbf{Left} describes a fact that the embedding distance between two procedures increases when the step-level difference denoted by Levenshtein distance gets larger. It is not surprising because the step order is preserved to some extent by all the methods, and large modifications in procedures will lead to distinct embedding differences. Thus, negative pairs with large step-level transformations will dominate the evaluation, which is unfair to those with small ones. To remedy this issue, WDR is introduced and it aims to evaluate embedding distance with respect to unit step-level transformation. Additionally, the curve in Figure \ref{fig5} \textbf{Right} proves the positive correlation between WDR and AUC. The proposed WDR is then expected to be a complementary measurement metric in this new task. 



\subsection{Scoring Demo}
\begin{figure}[t]
  \centering
  \includegraphics[width=0.95\columnwidth]{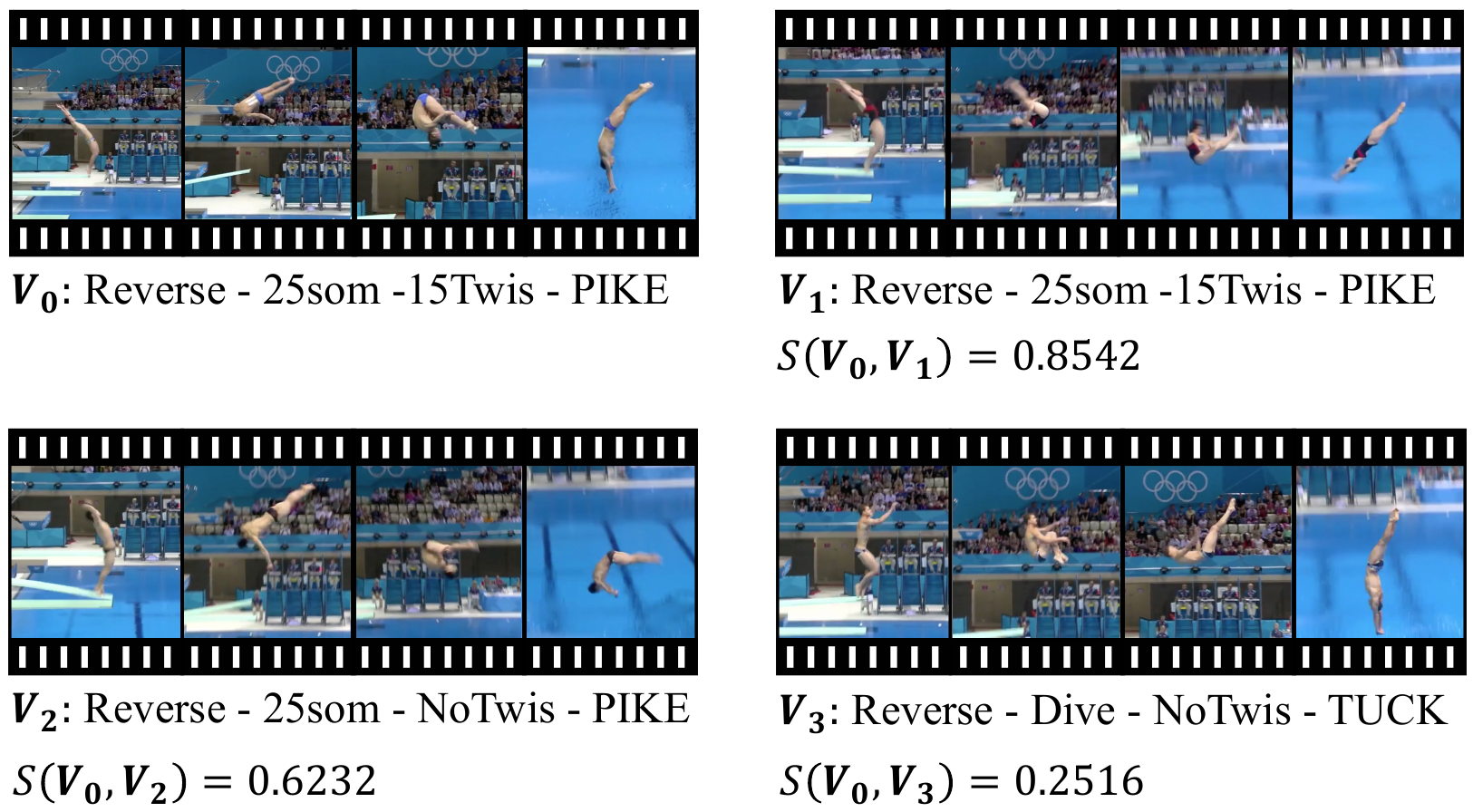}
  \caption{Videos with procedure annotation to be scored. $S(\cdot)$ is the scoring function, which calculates the cosine similarity between two videos in the embedding space.}
  \label{fig6} 
\end{figure}


As a potential solution to action assessment, sequence verification is able to act as a judge to score two procedures with fine-grained divergence. Here we show a diving scoring demo in Figure \ref{fig6}. The procedure in $V_0$ is chosen as the standard reference. We then calculate the cosine similarity as the score between the standard and each candidate video. One can easily tell that $V_1$ achieves the highest score since it performs the same procedure as $V_0$, while the scores of $V_2$ and $V_3$ decrease with the enlargement of their step-level difference compared to the standard. More demos are available in the supplementary material.

\subsection{Limitation and Impact}\label{sec5.7}
Though we have introduced two reorganized datasets and collected one scripted dataset for this new task, it still suffers from data insufficiency leading to an unsatisfactory performance on Video Swin Transformer. Also, it may prevent this promising task from being applied to the real application considering the generalization ability in the wild. The transformer encoder is introduced to aggregate temporal information, but it brings a parameter explosion as shown in Table \ref{table2}. 
Except for these, we hope this promising task could provide a novel insight for video understanding.  


\section{Conclusion}

In this work, we advocate a novel and interesting task sequence verification developed to verify two procedures with step-level differences when performing the same task. To that end, we reorganize two publicly available action-related datasets with step-procedure-task structure and collect an egocentric dataset asking volunteers to perform various scripted procedures. In addition to that, we develop a new evaluation metric that has been well verified to be a complement to the existing AUC metric. Finally, our proposed transformer-based method has been wildly studied and acts as a strong baseline for this new task. 

\section{Acknowledgements}
The work was supported by National Key R\&D Program of China (2018AAA0100704), NSFC \#61932020, \#62172279, Science and Technology Commission of Shanghai Municipality (Grant No.20ZR1436000), and 'Shuguang Program' supported by Shanghai Education Development Foundation and Shanghai Municipal Education Commission. 

\pagebreak
{\small
\bibliographystyle{ieee_fullname}
\bibliography{egbib}
}

\newpage
\appendix
\section{Appendix}
The appendix part contains two sections as the supplementary of the text, which is arranged as follows:
    
\textbf{1).} The first section contains some complementary information of our proposed CSV dataset, \textit{e.g.}, data gathering, annotations, and statistics information.

\textbf{2).} The second section gives more examples of scoring and a demo of another application, early warning.

\section{CSV Dataset}

The existing action datasets can hardly support our task due to the following reasons: i) some datasets focus on single actions and don't provide procedure videos; ii) some other datasets which contain procedure videos target other tasks such as action segmentation and action localization, \emph{i.e.}, they focus on the understanding of a single video rather than the verification of two videos, which leads to the lack of videos for similar procedures. However, the verification task indubitably requires a great number of videos that perform similar but slightly different step sequences for training. For the above reasons, we collect a new action veriﬁcation dataset to support our proposed task. In this section, we ﬁrstly describe the gathering process of the dataset, then give the annotation details of the videos, and finally demonstrate the statistical information of the dataset.

\subsection{Data Gathering}

\begin{figure}[t]
      \centering
      \includegraphics[width=0.95\columnwidth]{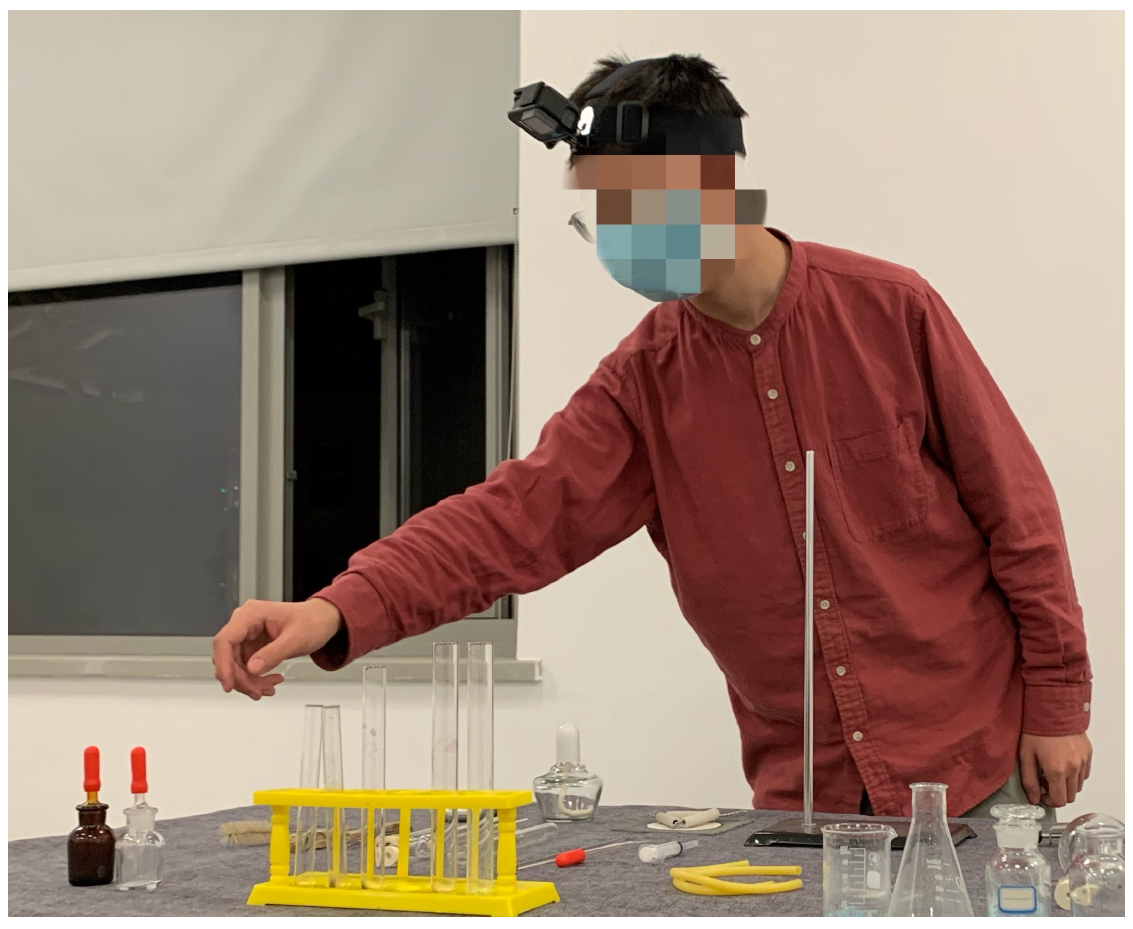}
      \caption{Head-mounted device used in data recording.}
      \label{Fig:device} 
    \end{figure}

    The dataset is recorded with the participation of 82 volunteers, whose ages range from 21 to 28, for performing scripted action sequences. Considering the constraints of venues, props, and personnel, we record videos of participants first setting up the equipment to perform a chemical experiment and then conducting that experiment. The specific process of recording is divided into the following steps: i) we firstly predefine 14 chemical experiment tasks, each of which contains consists of 5 procedures with a few step-level divergences, which will be detailed stated in Sec. \ref{annotation}; ii) the volunteers are required to remember these predefined operations and equip with a head-mounted camera (shown in Figure \ref{Fig:device}); iii) after the camera start working, the volunteers are asked to perform the predefined action sequences and put hands on the table or their sides when finished, and then the recording will be stopped. In this way, the integrity of procedures in the videos gets guaranteed.

    Following the collected method of \cite{damen2018scaling}, we choose GoPro HERO4 Black with an adjustable mounting such that the camera device can adjust to an appropriate pose with the variance of wearers' height, which provides multi-angle views and makes that each video contains interactions between the volunteers' hands and apparatus on the same experiment table. Besides, to ensure the stability and quality of the video, the camera is connected to a monitoring tablet via Bluetooth in order to monitor the quality of the recorded video at any time. Once a mistake occurs, the video will be discarded and re-shot. When shooting, the camera is set to the linear field of view, 24$fps$, and the resolution of 1920$\times$1080. Stereo audio is captured but discarded since almost all procedures proceed silently, and the sounds in videos will cause irrelevant noises. 

\subsection{Action Sequence Annotations}\label{annotation}

    In order to cater to our objective of verifying similar procedures with few step-level differences, we design fourteen different tasks in chemical experiments, and each enumerates all step-level transformations, \textit{e.g.}, additions, deletions, order exchange of steps. A step is defined as an action-object interaction whose label is always a combination of a verb, a noun, and sometimes prepositions. We label all procedures as $1.1 \sim 1.5,~2.1\sim2.5,~\cdots,~14.1\sim14.5$, totally 14 tasks, 70 labels. Note that we annotate each video only with a single serial number indicating the category of the procedure in the video, but without any temporal annotations, including the start and end frame of steps. Take the first task of procedures which is about screwing the test tube onto the iron stand and pouring water into the test tube as an example.

\begin{figure}[t]
    \centering
    \includegraphics[width=0.95\columnwidth]{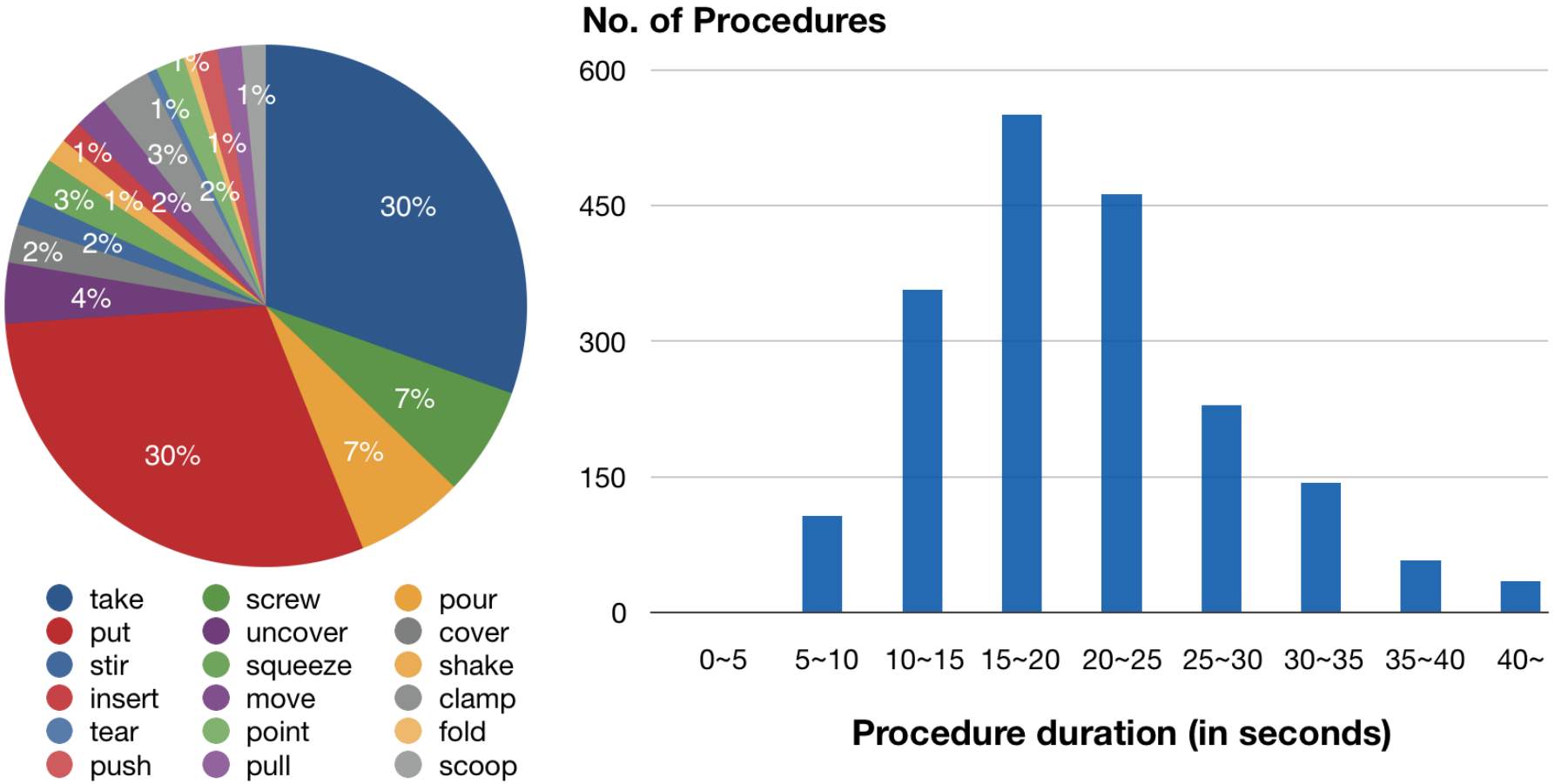}
    \caption{\textbf{Left}: Percentage of occurrences of each atomic-level action; \textbf{Right}: The histogram of step durations.}
    \label{Fig:CSV_statistics1} 
\end{figure}


\begin{itemize}
	\item \emph{1.1}: \textbf{take} (up the iron clamp) - \textbf{screw} (the iron clamp) - \textbf{take} (up the test tube) - \textbf{screw} (the iron clamp) - \textbf{take} (up the conical flask) - \textbf{pour} (the conical flask) - \textbf{put} (down the conical flask)
	\item \emph{1.2}: \textbf{take} (up the iron clamp) - \textbf{take} (the a test tube) - \textbf{screw} (the iron clamp) - \textbf{screw} (the iron clamp) - \textbf{take} (up the conical flask) - \textbf{pour} (the conical flask) - \textbf{put} (down the conical flask)
	\item \emph{1.3}: \textbf{take} (up the test tube) - \textbf{take} (up the conical flask) - \textbf{pour} (the conical flask) - \textbf{put} (down the conical flask) - \textbf{take} (up the iron clamp) - \textbf{screw} (the iron clamp) - \textbf{screw} (the iron clamp)
	\item \emph{1.4}: \textbf{take} (up the iron clamp) - \textbf{screw} (the iron clamp) - \textbf{take} (up the conical flask) - \textbf{put} (down the conical flask) - \textbf{take} (up the test tube) - \textbf{screw} (the iron clamp)
	\item \emph{1.5}: \textbf{take} (up the iron clamp) - \textbf{screw} (the iron clamp) - \textbf{take} (up the test tube) - \textbf{screw} (the iron clamp) - \textbf{take} (up the conical flask) - \textbf{put} (down the conical flask) - \textbf{take} (up the conical flask) - \textbf{pour} (the conical flask) - \textbf{put} (the conical flask)
\end{itemize}

    As illustrated above, compared to the \emph{1.1}, \emph{1.2} and \emph{1.3} disturb the order of actions; \emph{1.4} not only changes the order, but also deletes the \textbf{pour} action; and for \emph{1.5}, it inserts \textbf{take} - \textbf{put} actions into the standard one.

    The first group of procedures, which is a microcosm of the whole dataset, shows that most procedures differ in step order. The reason that we are so concerned about the order is that most action sequences will be unmeaning, sometimes even dangerous, if the order changes. For example, it is meaningless or even ridiculous to apply soap to hands after finishing washing hands.

\begin{figure*}[t]
    \centering
    \includegraphics[width=1.9\columnwidth]{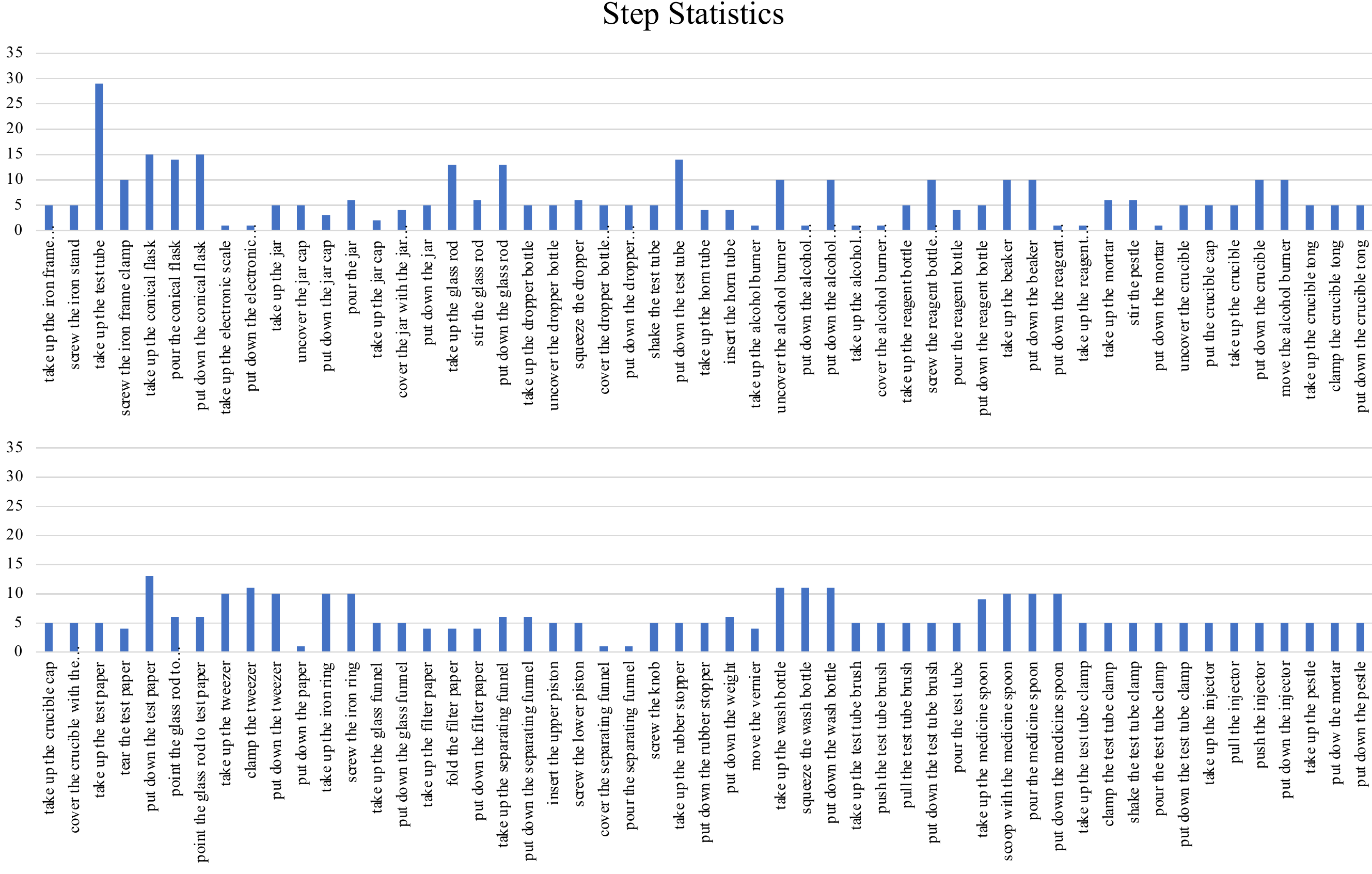}
    \caption{Step statistics. We list all 106 steps with their frequency of occurrence.}
    \label{Fig:CSV_statistics2} 
\end{figure*}

\subsection{Statistical Information}

    Figure \ref{Fig:CSV_statistics1} shows some statistics of our dataset.  As illustrated, we have 18 atomic-level actions with different frequencies in total, among which \textbf{take} and \textbf{put} are the two most common actions. This makes sense since taking up or putting down something is also extremely common in reality. By interacting one action with different objects, we have 106 steps in total (listed in Figure \ref{Fig:CSV_statistics2}). The videos' length varies from 5.63s to 58.43s due to the diversity in complexity among procedures and individual differences of participants, such as movement habits, the memory of the action sequence as well as familiarity with the operations. Totally, we collect 960,458 images of over 1,941 videos across 70 different kinds of procedures. On average, each video lasts 20.58 seconds, contains 495.85 frames, and consists of 9.53 steps.

\begin{figure*}[t]
  \centering
  \includegraphics[width=1.9\columnwidth]{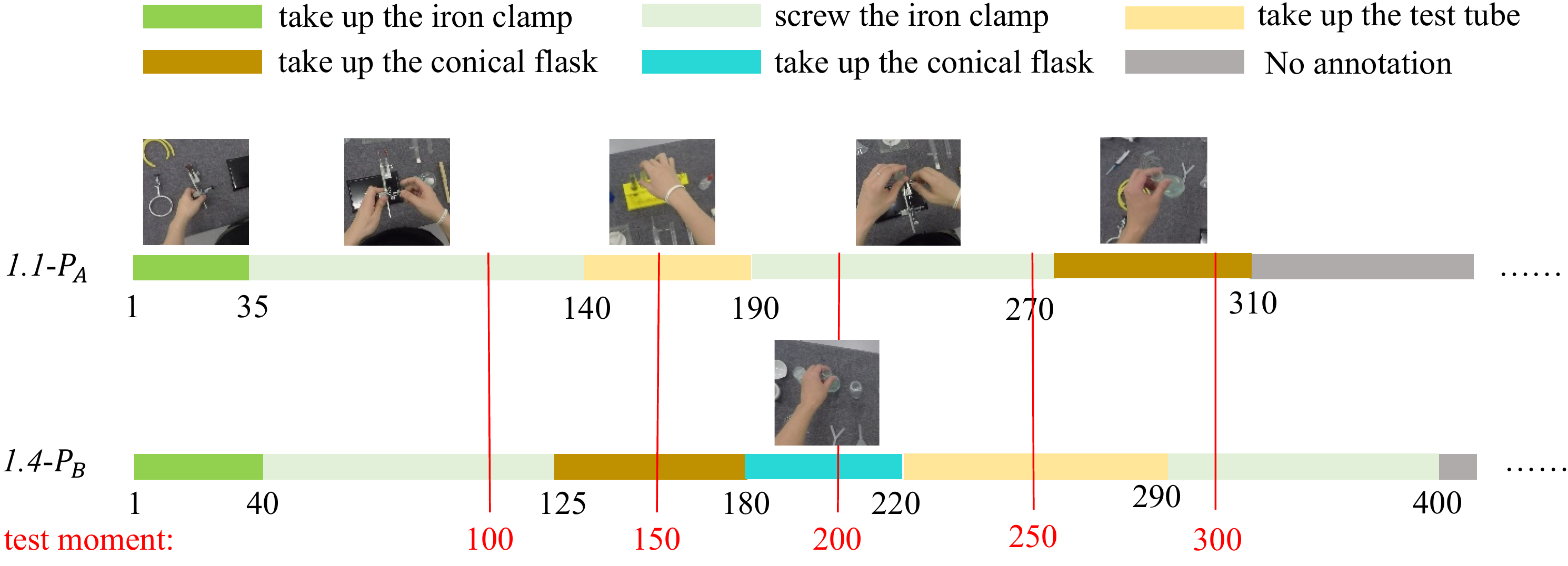}
  \caption{The temporal annotations of the exampled procedures. Blocks in the same color means that the corresponding clips of frames are annotated by the same step. The numbers on both sides of the block are the index of start and end frames of this action.}
  \label{Fig:CSV_annotation} 
\end{figure*}

\section{Demos}
\subsection{Scoring}

In this section, we demonstrate more examples as the scoring demo, which is detailed in Section 5.6 of the main body of this paper. For each dataset, we show two positive and two negative pairs, a total of eight videos with their procedure label. We can find Figure \ref{Fig:COIN_demo} has different procedure annotation from Figure \ref{Fig:diving_demo} and \ref{Fig:CSV_demo}, since the original COIN dataset \cite{tang2019coin} has temporal annotation for each step but Diving48 \cite{li2018resound} and CSV doesn't. It is worth noticing that $V_3$ and $V_4$ in Figure \ref{Fig:diving_demo} perform the same diving sequence but recorded from different directions of the athlete but still outputs a high matching score. 

\begin{figure}[t]
  \centering
  \includegraphics[width=0.95\columnwidth]{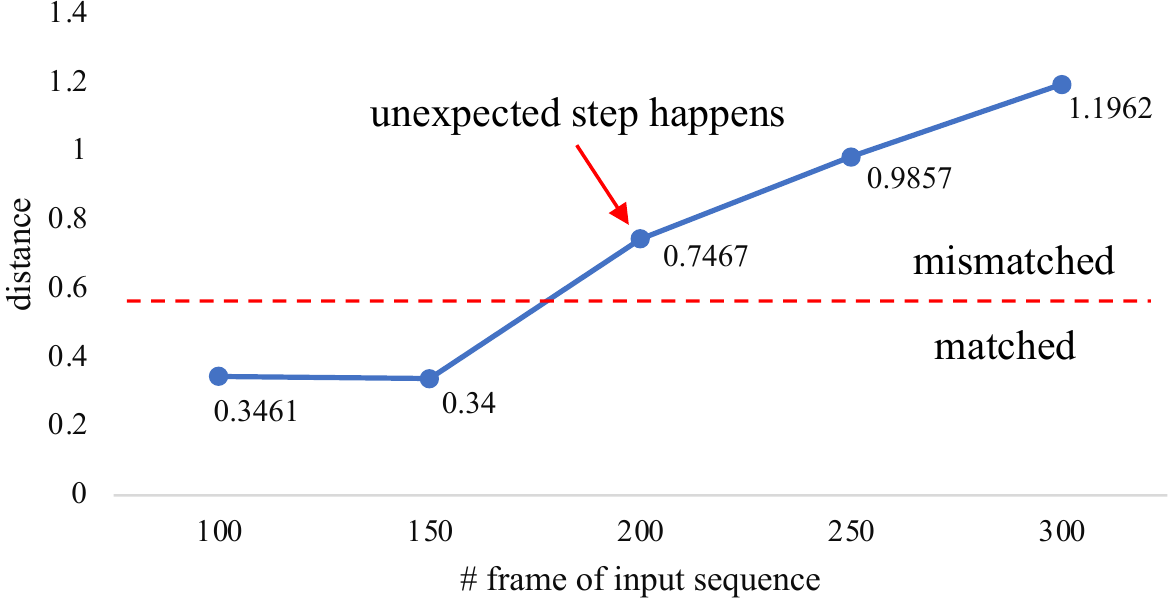}
  \caption{The evaluation result of the exampled video pair with on-line verification baseline.}
  \label{Fig:online_verification_curve} 
\end{figure}

\begin{figure*}[t]
    \centering
    \includegraphics[width=1.9\columnwidth]{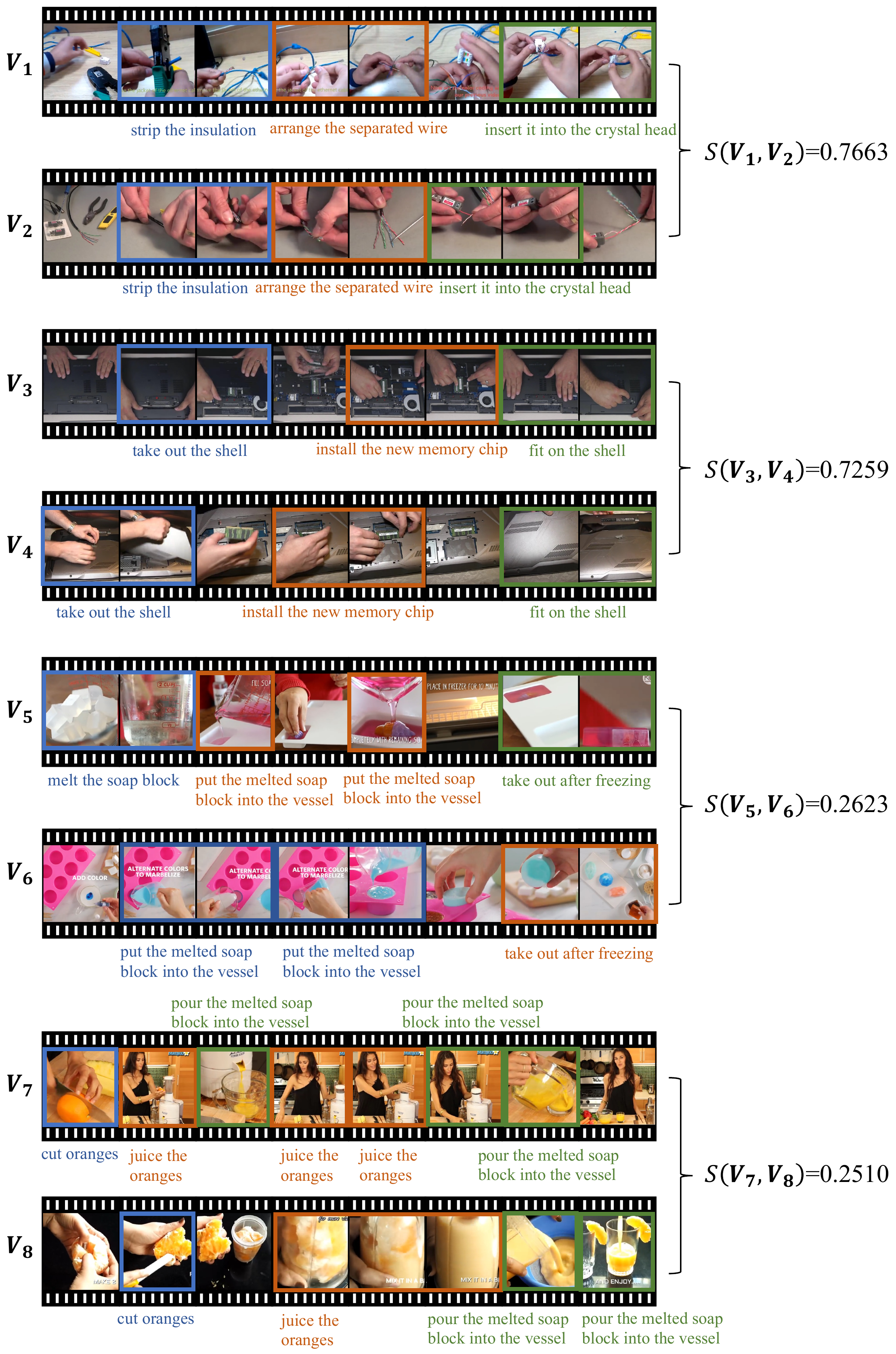}
    \caption{COIN-SV scoring example.}
    \label{Fig:COIN_demo} 
\end{figure*}

\begin{figure*}[t]
    \centering
    \includegraphics[width=1.9\columnwidth]{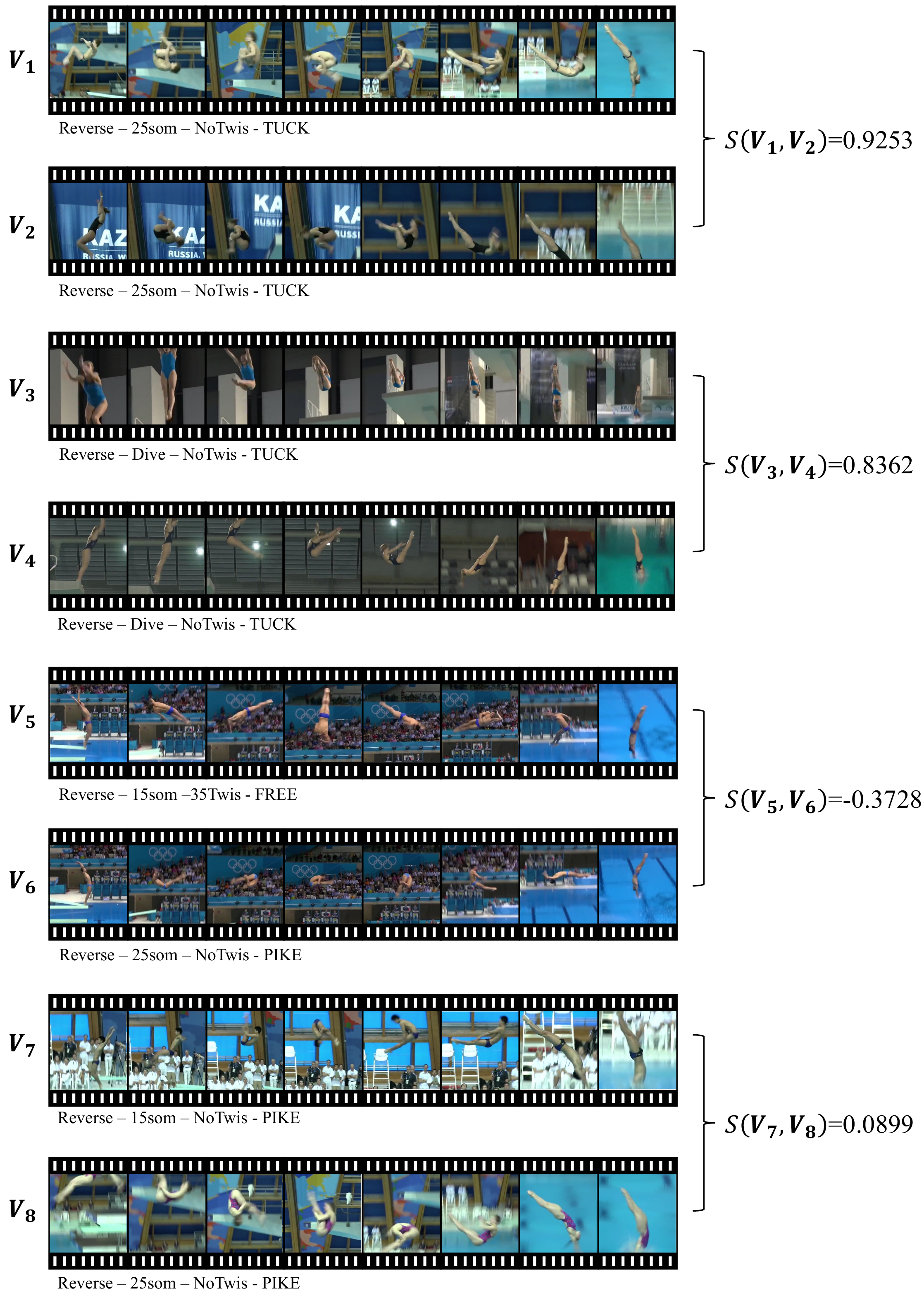}
    \caption{Diving48-SV scoring example.}
    \label{Fig:diving_demo} 
\end{figure*}

\begin{figure*}[t]
    \centering
    \includegraphics[width=1.9\columnwidth]{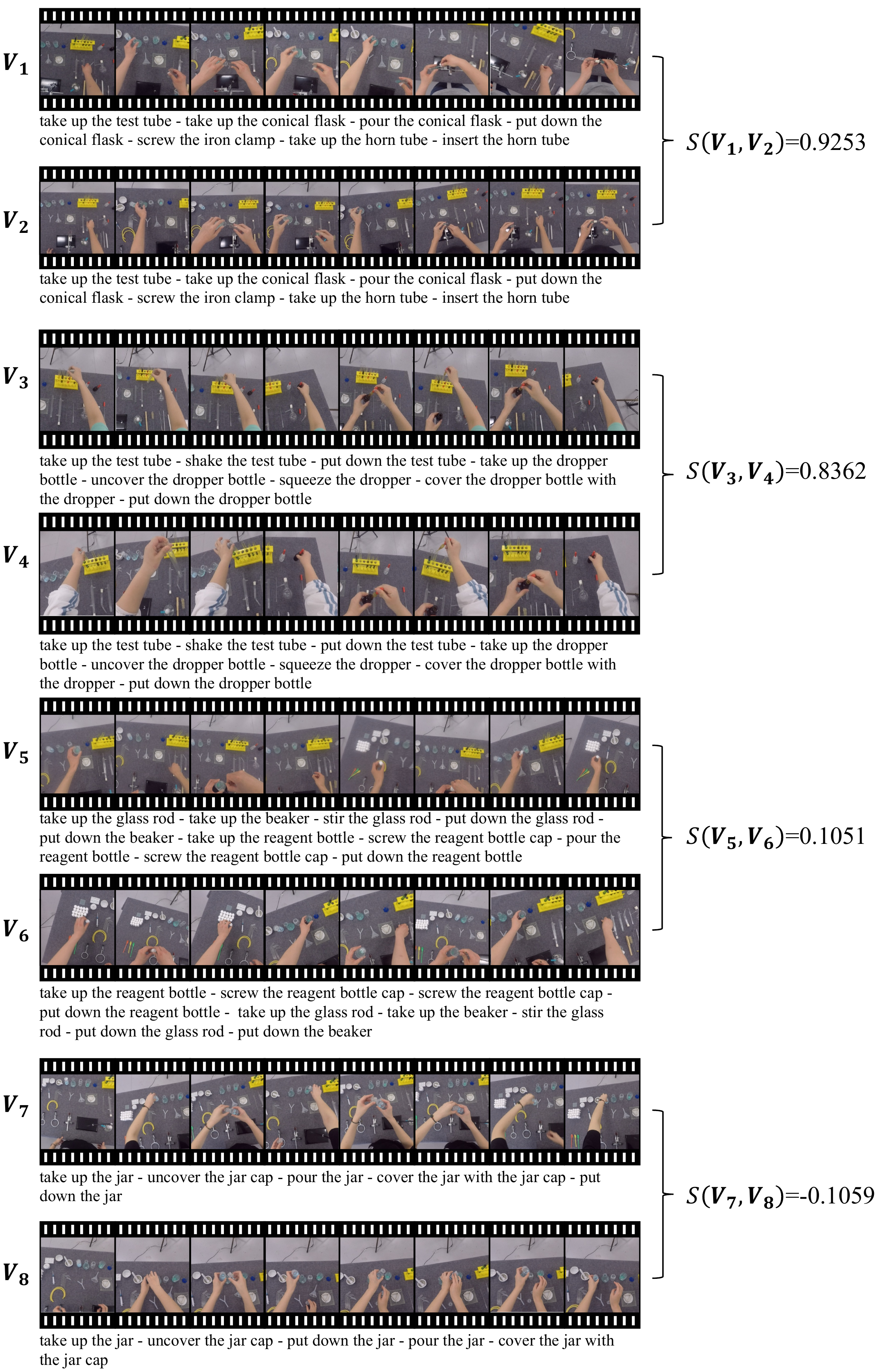}
    \caption{CSV scoring example.}
    \label{Fig:CSV_demo} 
\end{figure*}

\subsection{Early Warning}

    In addition to scoring, the sequence verification task can also be applied in early warning. The system is required to alarm whenever it detects the occurrence of an unexpected step. Thus, how to detect atomic-level actions in real-time and how to compare the incomplete input procedure with the complete reference procedure would be the main difficulties of this promising task, which is also our future research direction.

    
    However, the main body of this paper is to solve the verification problem of two complete procedures, which we named off-line verification. Here, we simply extend it to on-line sequence verification, where we can verify whether the input procedure is consistent with the reference in an on-line video stream. We design the following baseline. We take videos with labels \emph{1.1} and \emph{1.4}, which are performed by two participants $P_A$, $P_B$, for demonstration. According to the detailed illustration in Section \ref{annotation}, sequence \emph{1.4} and sequence \emph{1.1} are the same in the first three steps but are different in the fourth step. Note that although the third step are the same, the objects they interact with are different. However, such differences may be difficult for the model to recognize due to the light transmittance of glass products. The following is the specific description of the on-line action verification baseline.
    
    Given a $t$-frame test procedure $P_\text{test}$ and the corresponding reference procedure $P_0$, and assume that it takes similar time intervals for each individual to perform the same step (this assumption is the basis of the baseline). Then we can assume that $P_0[1:t\pm k]$ (the first $t\pm k$ frames of the reference procedure) is expected to perform the same step-sequence as $P_\text{test}[1:t]$ does if they are labeled the same, where $k$ is the time window size ($k=30$ in our experiment). For each $P_0[1:t+i], -k \leq i \leq k$, we calculate the $l_2$ distance between $P_0[1:t+i]$ and $P_\text{test}[1:t]$ in the feature space $f$ and average them over $2k+1$ cases as followed:
    \[\frac{\sum_{i=-k}^k\|f(P_\text{test}[1:t])-f(P_0[1:t+i])\|_2^2}{2k+1}\]

    Specifically, we stipulate all the frames of procedure \emph{1.1} performed by $P_A$ as the complete reference procedure, and the first 100/150/200/250/300 frames of procedure \emph{1.4} performed by $P_B$ as incomplete test procedures, the temporal annotation of these frames are given in Figure \ref{Fig:CSV_annotation}.
    
    Figure \ref{Fig:online_verification_curve} shows our experimental results. The blue line represents for the calculated $l_2$ distance in the feature space $f$ between \emph{1.4}-$P_B$ and \emph{1.1}-$P_A$ with different number of input frames. For the convenience of explanation, we notate the number of input test frames as $i$. When $i=100$, the value of \emph{distance} remains relatively low. This is because both the first 100 frames of \emph{1.4}-$P_B$ and the similar amount of frames of \emph{1.1}-$P_A$ perform the same steps. When $i=150$, note that although the objects interacted by the third step \textbf{take} around frame 150 are different in \emph{1.4}-$P_B$ and \emph{1.1}-$P_A$ (conical flask and test tube), such glass products are hard to distinguish by the model, which also leads to the small value of \emph{distance}. When $i=200$, the step in \emph{1.4}-$P_B$ is significantly different from the step in \emph{1.1}-$P_A$. Thus, the value of \emph{distance} rises rapidly. Besides, the broken line goes higher when $i=250$ or $300$ since more unmatched steps are included. We can easily catch the unexpected step in an on-line video stream through the huge jump of the line.

    According to above, when we choose an appropriate threshold of \emph{distance}, the \emph{1.4}-$P_B$ \emph{vs.} \emph{1.1}-$P_A$ pair is verified until the number of input frames achieves 200, the moment when the unmatched step occurs, which satisfies the requirement of on-line action verification. This section states a coarse mechanism for on-line action verification and evaluates a toy sample based on that, which can be applied in the field of early warning. We hope that this brick cast away can attract a jode, \emph{i.e.}, makes more researchers study this challenging but promising task.

\end{document}